\documentclass[conference,comsoc]{IEEEtran}
\usepackage{cite}

\ifCLASSINFOpdf
  \usepackage[pdftex]{graphicx}
\else
  \usepackage[dvips]{graphicx}
\fi
\usepackage[cmex10]{amsmath}
\usepackage{algorithmic}
\usepackage{array}
\ifCLASSOPTIONcompsoc
  \usepackage[caption=false,font=normalsize,labelfont=sf,textfont=sf]{subfig}
\else
  \usepackage[caption=false,font=footnotesize]{subfig}
\fi
\usepackage{dblfloatfix}
\usepackage{booktabs}
\usepackage{url}
\usepackage{float}
\usepackage{multirow}
\usepackage{enumerate}
\usepackage{amsthm}
\usepackage{nidanfloat}
\usepackage{threeparttable}

\newtheoremstyle{mystyle}
  {}
  {}
  {\itshape}
  {}
  {\bfseries}
  {.}
  { }
  {}
\theoremstyle{mystyle}

\newenvironment{talign*}
 {\let\displaystyle\textstyle\csname align*\endcsname}
 {\endalign}
\usepackage{arydshln}
\usepackage[colorinlistoftodos]{todonotes}
\usepackage[normalem]{ulem}
\usepackage{hyperref}
\usepackage{amsmath}
\DeclareMathOperator*{\argmax}{argmax} 
\usepackage{xurl}
\usepackage{amsmath}

\begin{document}

%
\title{Exploring Adversarial Attacks on Neural Networks: An Explainable Approach\\
}

\author{%
    Justus Renkhoff$^{~*1a}$, Wenkai Tan$^{~*1a}$, Alvaro Velasquez$^{2b}$, William Yichen Wang$^{5g}$, Yongxin Liu $^{1c}$, Jian Wang$^{3d}$\\ Shuteng Niu$^{4e}$, Lejla Begic Fazlic$^{6f}$,   Guido Dartmann$^{6f}$, Houbing Song$^{1c}$\\
    $^{1}$Embry-Riddle Aeronautical University, FL 32114 USA,
    $^{2}$University of Colorado Boulder, CO 80309 USA\\
    $^{3}$University of Tennessee at Martin, TN 38237 USA,
    $^{4}$Bowling Green State University, OH 43403 USA, \\
    $^{5}$Purdue University, IN 47907 USA,
    $^{6}$Trier University of Applied Sciences, Germany\\

    $^{a}$\{renkhofj, tanw1\}@my.erau.edu,
    $^{b}$alvaro.velasquez.1@us.af.mil,
    $^{c}$\{LIUY11, songh4\}@erau.edu,\\
    $^{d}$jwang186@utm.edu, $^{e}$sniu@bgsu.edu
    $^{f}$\{l.begic, g.dartmann\}@umwelt-campus.de,
     $^{g}$ wywang@purdue.edu\\
    \thanks{* Wenkai Tan and Justus Renkhoff are co-first authors.}
}

\markboth{IEEE Internet of Things Journal,~Vol.~11, No.~4, May~2021}%
{Shell \MakeLowercase{\textit{et al.}}: Bare Demo of IEEEtran.cls for Journals}
\IEEEtitleabstractindextext{%
\begin{abstract}

Deep Learning (DL) is being applied in various domains, especially in safety-critical applications such as autonomous driving. Consequently, it is of great significance to ensure the robustness of these methods and thus counteract uncertain behaviors caused by adversarial attacks. In this paper, we use gradient heatmaps to analyze the response characteristics of the VGG-16 model when the input images are mixed with adversarial noise and statistically similar Gaussian random noise. In particular, we compare the network response layer by layer to determine where errors occurred. Several interesting findings are derived. First, compared to Gaussian random noise, intentionally generated adversarial noise causes severe behavior deviation by distracting the area of concentration in the networks. Second, in many cases, adversarial examples only need to compromise a few intermediate blocks to mislead the final decision. Third, our experiments revealed that specific blocks are more vulnerable and easier to exploit by adversarial examples. Finally, we demonstrate that the layers $Block4\_conv1$ and $Block5\_cov1$ of the VGG-16 model are more susceptible to adversarial attacks. Our work could potentially provide useful insights into developing more reliable Deep Neural Network (DNN) models.

\end{abstract}

}

\IEEEoverridecommandlockouts

\maketitle

\IEEEdisplaynontitleabstractindextext

%
\IEEEpeerreviewmaketitle

\section{Introduction}
%
%
%
%
Artificial intelligence (AI) and deep learning (DL) provide unlimited possibilities for addressing various engineering and scientific problems. However, the reliability and robustness of Deep Neural Networks (DNNs) has caused many concerns; for example, some researchers revealed that DNNs, such as the VGG-16 model \cite{simonyan2014very}, can be misled by intentionally mutated images that are imperceptible to humans\cite{10.1145/3132747.3132785,miller2020adversarial,luo2018towards}. In these scenarios, the mutated pixels have pseudo-random characteristics and thus raise concerns about the uncertainty and trustworthiness of DNNs under the natural Gaussian noise of their operational environments \cite{englert2019machine,ignatiev2020towards}.

For mitigation, on the one hand, some solutions are proposed to increase the robustness of DNNs by augmenting their training with perturbed samples or introducing a robust loss term \cite{wu2021wider,gong2021maxup,gao2019convergence}. These approaches encourage DNNs to treat a slightly perturbed image as its origin. In this context, finding and incrementally training DNNs with adversarial examples is analogous to fuzzy testing. Some representative frameworks are proposed, such as DLFuzz \cite{9099600}, DeepXplore \cite{10.1145/3132747.3132785}, DeepHunter \cite{xie2019deephunter}, and TensorFuzz \cite{odena2019tensorfuzz}. 

One common feature of these adversarial example-enabled neural network fuzzy testing frameworks is that they not only discover adversarial examples but also try to maximize the activation rates of neurons, a.k.a. neuron coverage \cite{yang2022revisiting, harel2020neuron}. Neuron coverage describes how many neurons are activated during a prediction. DLFuzz \cite{9099600} adapts this concept from DeepXplore \cite{10.1145/3132747.3132785} and tries to optimize this metric by generating adversarial examples and maximizing the prediction difference between the original and the adversarial images. Higher neuron coverage usually contributes positively to the robustness of DNNs.

However, training DNNs on perturbed samples incrementally is computationally expensive and can reduce classification accuracy \cite{lamb2019interpolated}. Moreover, it is difficult to find a balance between accuracy and adversarial robustness. Defending existing DNNs against adversarial examples is preferable. Defense-GAN \cite{samangouei2018defense} trains a defensive generative adversary network (GAN) on natural inputs. A noticeable behavior deviation can be detected when adversarial examples are fed into the defensive GAN. Instead of modeling the inputs directly, I-Defender \cite{zheng2018robust} models the output distributions of fully connected hidden layers on each class. Then it uses statistical testing to reject adversarial examples. Adversarial perturbations can be treated as additive noise. Therefore, similar approaches, such as denoising autoencoders, can be used to purify the input of DNNs \cite{gondara2016medical, gu2014towards, yadav2022integrated,hwang2019puvae}.

Most of the current efforts still regard DNNs as black-box models and have not yet analyzed the effect of adversarial attacks in an explainable way. In this paper, we use images from the ImageNet database \cite{5206848} and then manipulate them with DLFuzz \cite{9099600}, which generates adversarial examples based on given seed images and tries to activate as many neurons as possible simultaneously. Grad-CAM \cite{Selvaraju_2017_ICCV} heatmaps are generated to make the decision-making procedure explainable. Wrongly classified mutated images are analyzed and compared to their origin to find out why and in which layer of the behavior deviations occur. We compare the response characteristics of the VGG-16 model using Grad-CAM when the input images are mixed with adversarial and statistically similar Gaussian noise. Our findings are as follows.
\begin{itemize}
    \item Both random noise and adversarial noise cause behavioral deviations of intermediate layers. However, adversarial noise causes more behavioral deviations by distracting the area of focus in the intermediate layers.
    \item There are certain blocks that are more vulnerable and more easily exploited by adversarial examples. In particular, we demonstrate that the layers $Block4\_conv1$ and $Block5\_cov1$ of the VGG-16 model are more susceptible to adversarial attacks.
    \item We show that a neural network model can be misled by compromising only a few blocks.
\end{itemize}

The remainder of this paper is organized as follows: A literature review of related work is presented in Section~\ref{sectRW}. We present the methodology in Section~\ref{sectMM}. The evaluation and discussion are presented in Section~\ref{sectEED} with the conclusions in Section~\ref{sectCC}.

\section{Related Work}
\label{sectRW}
Current efforts aim at increasing the robustness of DNNs against adversarial attacks. Accordingly, current research addresses approaches to make the decision-making of DNNs more transparent \cite{8591457} \cite{selvaraju2017grad} \cite{ribeiro2016should}, and to find vulnerabilities in model architectures \cite{lsa} \cite{Haizhong}. Since this research is utilized and closely related to our approach, we discuss these efforts in the following.

\textit{Adversarial Attacks}: An adversarial example is generated by introducing small and imperceptible perturbations to a given seed image to cause misclassifications. There are several procedures, such as FGSM \cite{goodfellow2014explaining}, to generate adversarial examples. In general, an adversary attacker has to know the parameters of the target neural classifier and then solve an optimization problem that the perturbations should maximize the classification loss and minimize the difference between perturbed and original image. Defending DNNs against adversarial examples can increase the robustness of the model.

\textit{Visual Explanations}: Methods like Gradient-weighted Class Activation Mapping (Grad-CAM) \cite{selvaraju2017grad} and Local Interpretable Model-Agnostic Explanations (LIME) \cite{ribeiro2016should} aim to improve the interpretability of DNNs. They have the ability to explain the prediction of black-box models, and can therefore improve their trustworthiness. Grad-CAM calculates a heatmap that highlights different areas of an image in different colors. These colors visualize how much an area positively contributes to a certain prediction. LIME highlights pixels that contribute positively or negatively on the basis of a threshold. These pixels can be represented in different colors, as in Grad-CAM. With Grad-CAM, it is possible to identify not only which areas contribute positively and negatively to a certain prediction, but also how much an area influences a decision \cite{cian2020evaluating}. There are numerous efforts that use or improve these methods to increase confidence in machine learning models \cite{2008.02312} \cite{poerner-etal-2018-evaluating} \cite{app112110417}.

\section{Methodology}
\label{sectMM}
\subsection{Problem Formulation}
\begin{figure}[h]
    \centering
    \includegraphics[width=0.9\linewidth]{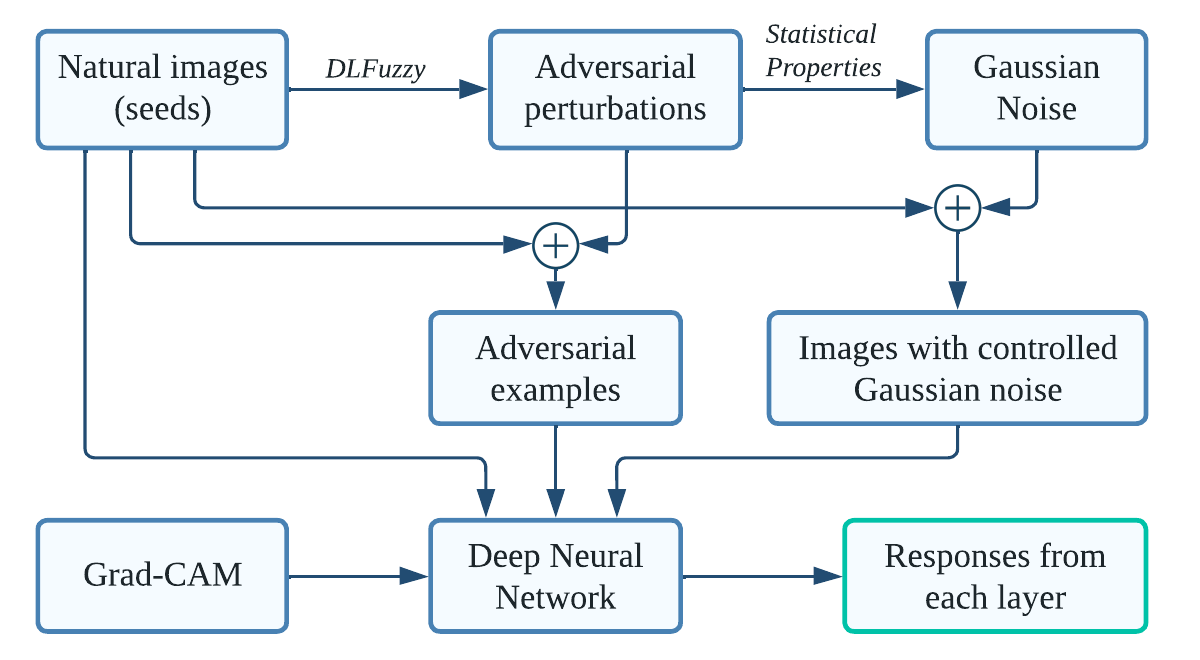}
    \caption{Exploring neural network response against adversarial perturbations and Gaussian noise.}
    \label{figWorkFlow}
\end{figure}

To make DNNs as robust as possible to adversarial attacks, we must understand how adversarial examples cause misclassifications. For this reason, the classification process is analyzed layer by layer to find sections of an exemplary selected DNN that are particularly vulnerable to such attacks. Herewith, we want to present a procedure to analyze any DNN, making it possible to find vulnerabilities in the fundamental network architecture to be able to counteract these in the future. A brief workflow of this paper is given in Figure~\ref{figWorkFlow}. 

In general, original samples (a.k.a. natural images), adversarial examples, and noisy samples are then analyzed using Grad-CAM. A heatmap is generated for each layer of the target VGG-16 model showing what the focus point of the DNN in the corresponding layer is. Then, the cosine similarity is calculated between the heatmaps of the adversarial examples and those of the original images for every layer. By locating layers where the similarity between the adversarial and the original samples is particularly low, we can determine which layers react strongly to the perturbations.
\subsection{Data Preparation}
We used randomly selected images as seeds from the ImageNet dataset, and we derive their adversarial perturbations (a.k.a. adversarial noise), denoted as:
\begin{align}
\label{eqADVPerturb}
    \mathbf{N}_a = \mathrm{ADV}(\mathbf{S}, \varepsilon)
\end{align}
where $\mathrm{ADV}(\cdot)$ is the adversarial perturbation function, $\varepsilon$ is the strength of the perturbation, and $\mathbf{S}$ denotes a seed image.  In this work, DLFuzz is configured to generate adversarial noise and maximize neuron coverage at the same time. We generate the same amount of Gaussian noise $\mathbf{N}_g$, defined as random perturbations, with statistical properties similar to the adversarial perturbations. Accordingly, we use the expected value, standard deviation and shape of $\mathbf{N}_a$ to generate $\mathbf{N}_g$. We define the distribution as:
\begin{align}
\label{eqGaussianNoise}
    \mathcal{N}(\mu_a, \Sigma_a)    \,
\end{align}
where $\mathcal{N}(\cdot)$ denotes a normal distribution, $\mu_a$ denotes the expected value, and $\Sigma_a$ denotes the standard deviation. We let $\mu_a = \overline{\boldsymbol{N}_a}$ and $\Sigma_a = \sigma(\boldsymbol{N}_a)$. Based on this distribution, we generate random noise $\mathbf{N}_g$ represented as a matrix with the same shape as $\mathbf{N}_a$. For more details, please refer to our implementation in GitHub\footnote{\url{https://github.com/JustusRen/Exploring-Adversarial-Attacks-on-Neural-Networks}}.  
In this way, the augmented input $\mathbf{I}$ to the DNN becomes:
\begin{align}
    \mathbf{I} = \{\mathbf{S}, \mathbf{S}+\mathbf{N}_a, \mathbf{S}+\mathbf{N}_g\}
\end{align}
where $\mathbf{S}+\mathbf{N}_a$ and $\mathbf{S}+\mathbf{N}_g$ are adversarial examples and the noisy version of the seed image.

\subsection{Network Behavior Deviation Detection}
\label{sectBDD}
For a given convolutional layer $l$, its response to an input image, e.g., a seed image, can be visualized using its grad-CAM heatmap, defined as:
\begin{align}
    \mathbf{H(\mathbf{S})} = R \left[ sign \left( \dfrac{\mathbf{J(\mathbf{S})}}{\partial\mathbf{\theta_l}} \right) \right]
\end{align}
where $\mathbf{J(\mathbf{S})}$ denotes the classification loss given $S$ and $\mathbf{\theta_l}$ denotes the parameters of layer $l$. The function $R(\cdot)$ denotes the operation that reshapes and interpolates the derived gradients to the same dimension and size as $\mathbf{S}$. The Grad-CAM heatmap displays the focal areas of $l$ in $\mathbf{S}$. Therefore, the degree or level of behavioral deviations of the heatmap under the perturbed sample and the natural image can be used to quantify whether it is compromised. Hereby, we use cosine similarity to calculate the degree of behavioral deviations as:
\begin{align}
     \label{eqCosineSim}
     \mathbf{G}[\mathbf{S_1},\mathbf{S_2}] = \dfrac{vec(\mathbf{H(\mathbf{S_1}))} \cdot vec(\mathbf{H(\mathbf{S_2}))}}{\| vec(\mathbf{H(\mathbf{S_1}))} \|\cdot \|  vec(\mathbf{H(\mathbf{S_2}))} \|}
    \end{align}
where $\mathbf{H(S_1)}$ and $\mathbf{H(S_2)}$ denote Grad-CAM heatmaps of different images. These are vectorized and normalized for the similarity calculation. For a specific layer and a seed image $\mathbf{S}$, we calculate and compare two degrees of behavioral deviation:
\begin{align}
    \mathbf{D}_a = \mathbf{G}[\mathbf{S}, \mathbf{S}+\mathbf{N}_a]\\
    \mathbf{D}_g =\mathbf{G}[\mathbf{S}, \mathbf{S}+\mathbf{N}_g]
\end{align}

Our observations revealed that both Gaussian noise and adversarial perturbations can cause behavioral deviations. For each layer, we used the median value of its degree of behavioral deviation under Gaussian noise as the threshold to quantify whether it is compromised, i.e. whether its behavior deviates more severely than when it is under the same amount and strength of Gaussian noise.

\subsection{DLFuzz for Adversarial Example Generation}
We use DLFuzz \cite{9099600} to calculate perturbations and applied to images from the ImageNet \cite{ILSVRC15} data set. The procedure of deriving additive perturbations is regarded as an optimization problem as:
\begin{align}
    \argmax_{\mathbf{N}_a,~||\mathbf{N}_a|| \leq \delta}\left[ \sum^K_{i=1}c_i-c +\lambda\sum^m_{i=0}n_i\right]
\end{align}
where $\delta$ restricts the magnitude of the adversarial perturbation, $c$ is the prediction on a given seed image, $c_i$ is one of the top $K$ candidate predictions, $n_i$ is one of the activation values in the $m$ selected neurons, and $\lambda$ is a constant to balance between activating more neurons and obtaining adversarial examples.

Mathematically, DLFuzz manipulates the perturbations to mislead a tested network to make wrong predictions and simultaneously activate the selected neurons. By maximizing neuron coverage, DLFuzz ensures that adversarial examples are generated more efficiently during a test. Based on the assumption that this will trigger more logic in the network and thus provoke and detect more erroneous behavior \cite{10.1145/3132747.3132785}, the underlying DNN we selected is the VGG-16 \cite{simonyan2014very} model pre-trained on the ImageNet dataset. 

After obtaining an effective adversarial perturbation for each seed, we amplify the perturbations with a total of five different ratios, defined as the perturbation strength. These are 25\%, 50\%, 100\%, 200\% and 400\%.


\section{Evaluation and Discussion}
\label{sectEED}
In this section, we evaluate the behavior deviation of a pre-trained VGG-16 network under adversarial examples and Gaussian noise. 
\subsection{Seed Selection}
The ImageNet dataset contains more than 10 million images in 1,000 categories. We randomly pick five images from each category to derive their adversarial examples. Unfortunately, not all images can find their corresponding adversarial examples within the pre-defined magnitude of perturbation. Ultimately, we derived at least one adversarial example from 48\% of the randomly collected images.

\subsection{Neuron Coverage under different attacks}

\begin{figure}[h]
    \centering
    \subfloat[Neuron Coverage - random and deliberate perturbations]
    {%
        \includegraphics[width=0.8\linewidth]{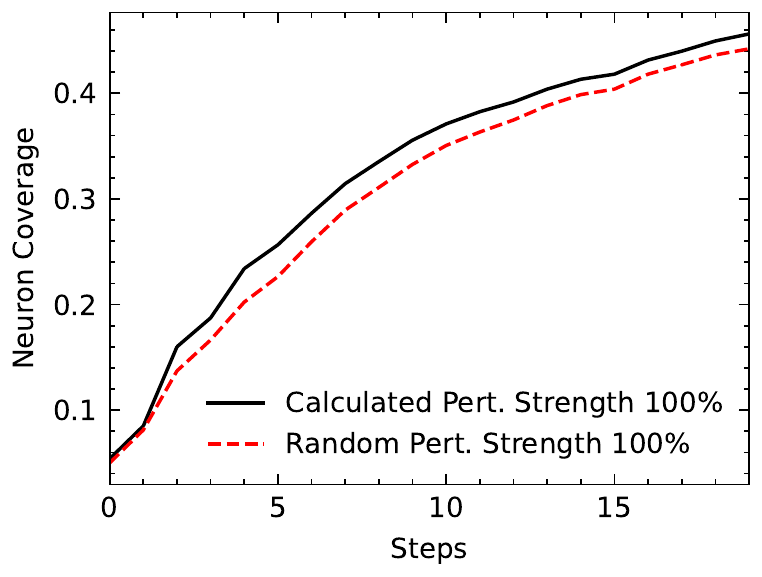}
        \label{figNeuronCoverage}
    }\\
    \subfloat[Neuron Coverage - different perturbation strengths]
    {
        \includegraphics[width=0.8\linewidth]{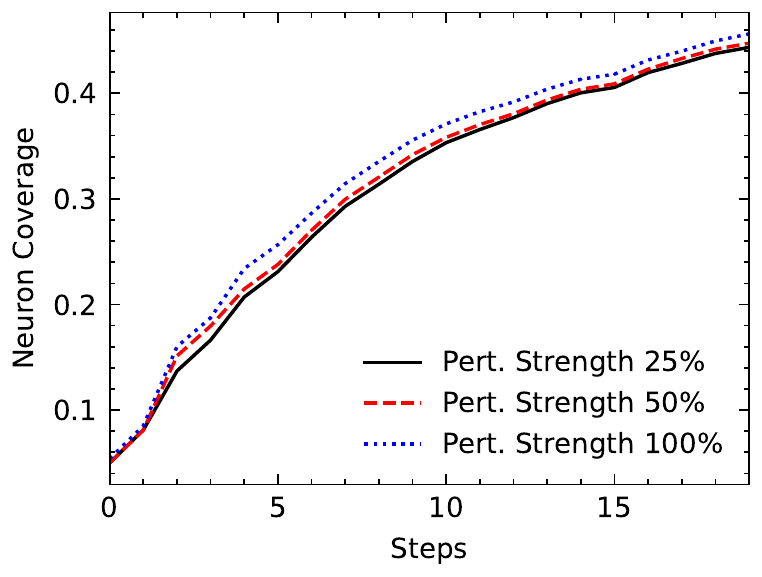}
        \label{figNeuronCoverage_perturb}
    }%
    \caption{Comparison of neuron coverage optimization with random perturbations, targeted perturbations and different perturbation strengths}
\end{figure}

Figure \ref{figNeuronCoverage} shows the progression of the neuron coverage of the model using images with adversarial noise and random noise. It is noticeable that the neuron coverage increases slightly faster using adversarial perturbations. However, DLFuzz can also increase neuron coverage using random perturbations.
Furthermore, Figure \ref{figNeuronCoverage_perturb} shows how the neuron coverage increases when different perturbations strengths are applied to the seed images. In general, the strength of the perturbation has a slight effect on the neuron coverage. The less severe the perturbation, the slower the neuron coverage increases. 


\subsection{Network Vulnerability Analysis}

\begin{figure*}
    \centering
    \includegraphics[width = \textwidth]{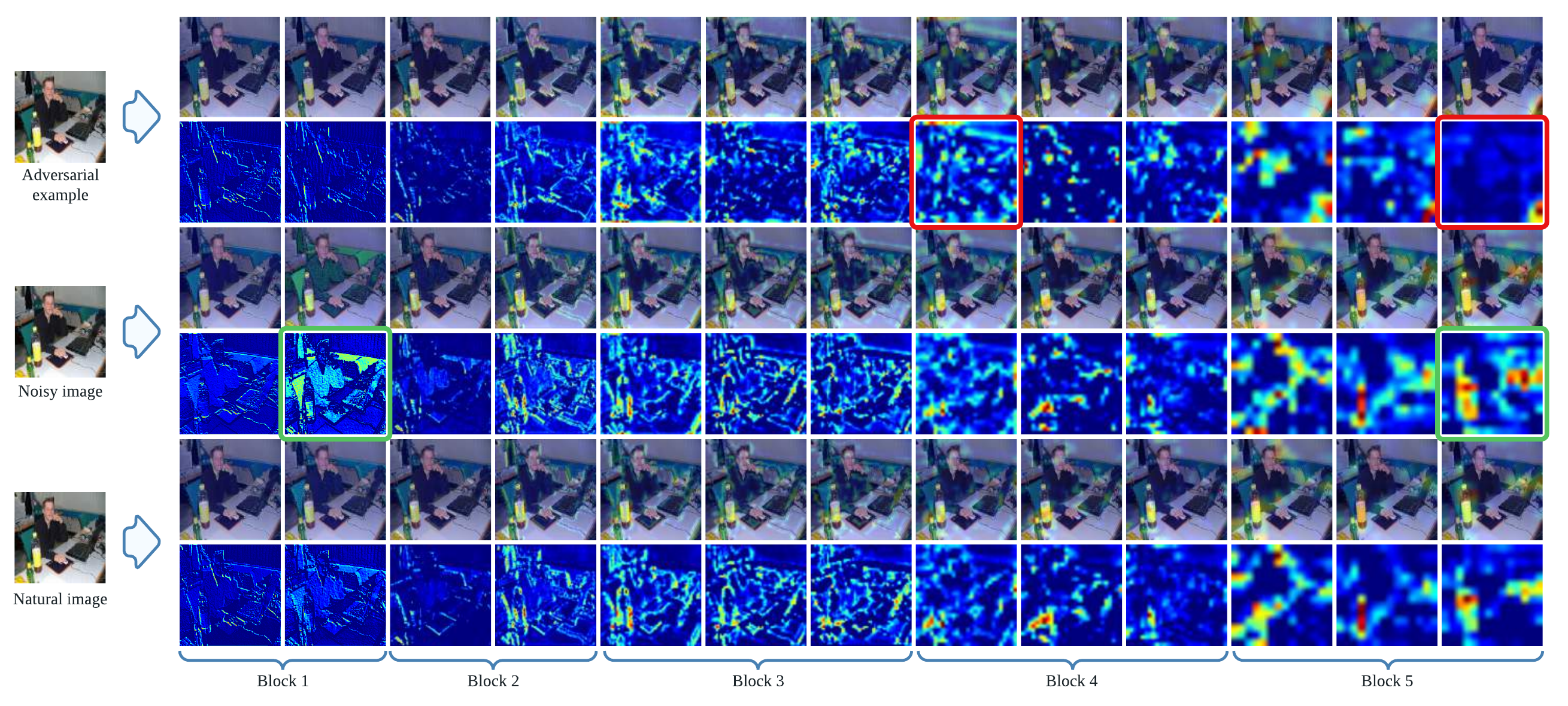}
    \caption{Gradient CAMs of VGG-16 under different inputs. Red and green rectangles highlight the convolutional layers that have significant behavior drifts.}
    \label{figCAMArray}
\end{figure*}

\begin{figure}[h]
\centering  
\subfloat[]
{%
    \includegraphics[width=0.75\linewidth]{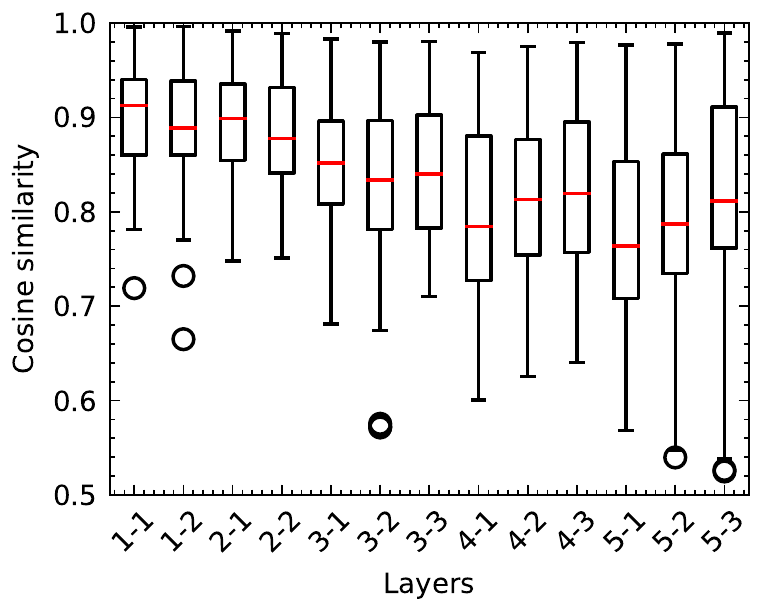}
    \label{figBoxlotNormal}
}\\
\subfloat[]
{
    \includegraphics[width=0.75\linewidth]{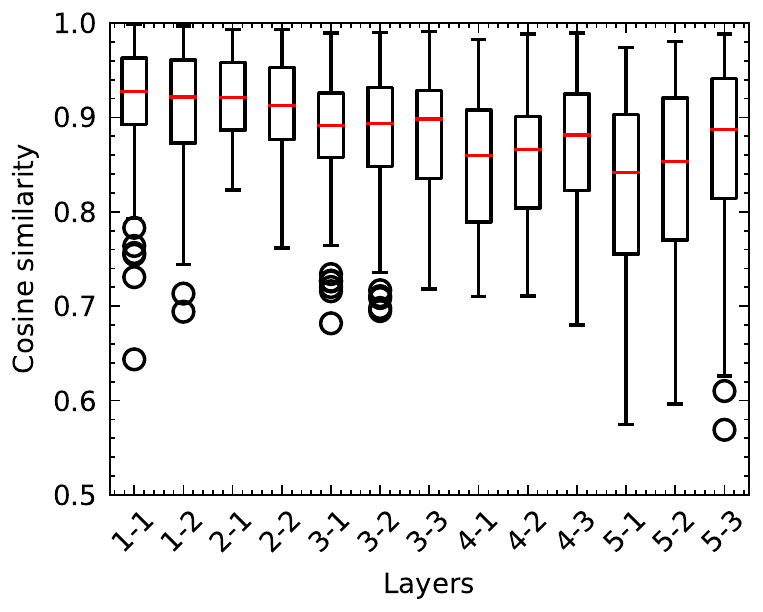}
    \label{figBoxplotRandom}
}%
\caption{Comparison of behavioral deviations under: (a) adversarial examples and (b) random noise for the VGG-16 convolutional layers (x\text{-}axis 1\text{-}1 represents the Block1\_Conv1 layer in VGG-16).}
\label{figCosSimBoxplot}
\end{figure}

We used the Grad-CAM heatmaps of each layer under natural images as references and used cosine similarities to measure their behavior deviations under different attacks as shown in Figure~\ref{figCAMArray}. More specific statistical analysis is presented in Figure~\ref{figBoxlotNormal}, where the Grad-CAM responses of the neural network under adversarial examples are severely deviated. More specific statistical analysis is presented in Figure \ref{figBoxlotNormal}, where the Grad-CAM responses of the neural network under adversarial examples are severely deviated. Under adversarial examples, each layer’s heatmap cosine similarities to the natural image have lower mean values and greater variances in comparison with the similarities under noisy images.

In particular, the two layers, Block4\_Conv1 and Block5\_Conv1, are considered to be easier to compromise because their response deviates more significantly under adversarial noise, showing a lower average cosine similarity than when it receives images with pure Gaussian noise. Interestingly, Gaussian noise also causes network behaviors to deviate from those when the network receives only clear images. As shown in Figure~\ref{figBoxplotRandom}, although we observe lower cosine similarities, the response deviation is not as significant as when the network is under adversarial examples.


\subsection{Behavior Drift Under Different Attack Strengths}
As we mentioned, Gaussian noise and adversarial perturbations can cause deviations in network behavior. Adversarial perturbations can drift the network behavior more and play a significant role in adversarial attacks. As given in Equation \ref{eqGaussianNoise}, the Gaussian random perturbations will have statistical properties consistent with adversarial perturbations.

\begin{figure}[h]
\centering  
\subfloat[]
{%
    \includegraphics[width=0.75\linewidth]{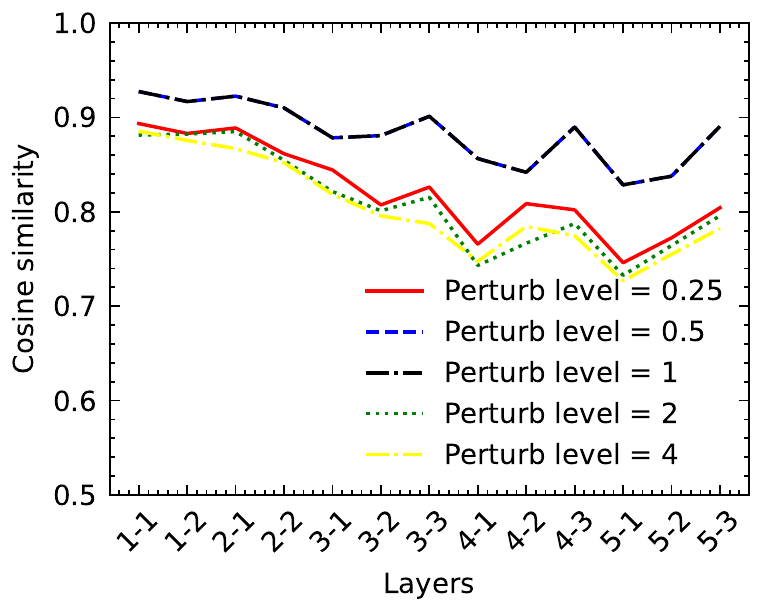}
    \label{figLineplotNormal}
}\\
\subfloat[]
{
    \includegraphics[width=0.75\linewidth]{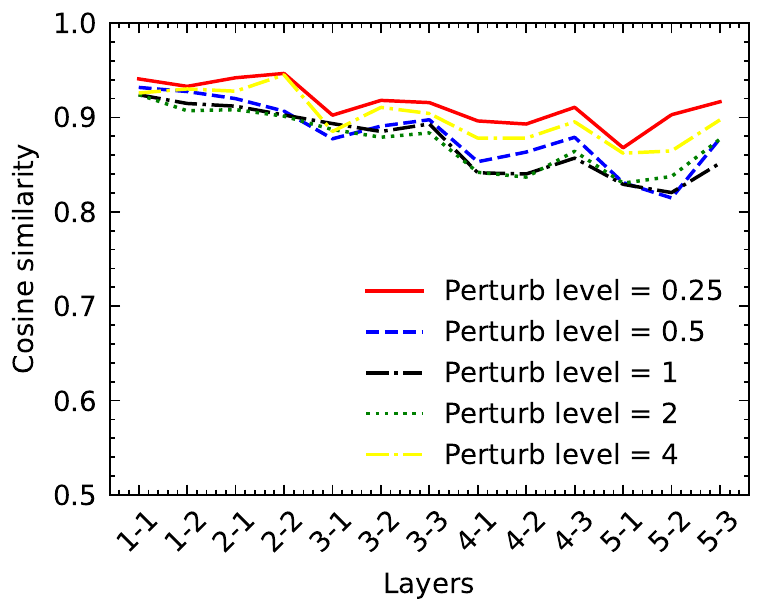}
    \label{figLineplotRandom}
}%
\caption{Comparison of average behavioral deviation on: (a) adversarial examples and (b) images with Gaussian noise with different attack strength (x\text{-}axis 1\text{-}1 represents the Block1\_Conv1 layer in VGG-16).}
\label{figCosSimLineplot}
\end{figure}

According to Figure~\ref{figLineplotNormal} and \ref{figLineplotRandom}, the comparison of cosine similarities in grad-CAM heatmaps indicates that adversarial inputs have a lower overall value (a.k.a., more severe behavior deviations) than noisy inputs at all attack levels. When we adjust the attack strength, the cosine similarities on adversarial inputs decrease more. The Grad-CAM heatmap cosine similarities decrease more significantly, indicating more severe behavioral deviations as determined by a smaller cosine similarity, when adversarial perturbations are amplified 2 and 4 times. Comparably, although the network's averaged behavior also deviates under Gaussian noisy images, the averaged magnitude of deviation is not as significant as under adversarial perturbations. Although it is generally believed that a stronger adversarial attack strength can cause the behavior of the neural network to deviate further, our experiments show that the behavioral deviation is still significant when the attack strength is reduced to 0.25, for which the three curves when the attack strength are 0.25, 2 and 4 are very similar. Interestingly, behavioral deviations under attack strength 0.5 and 1 are almost identical.

In general, more significant behavioral deviations can be observed in deeper layers of the network, as shown in Figures~\ref{figLineplotNormal} and \ref{figLineplotRandom}. However, we do notice some fluctuations, which indicates that there are some specific layers that are more vulnerable and easier to compromise.

\subsection{Distributions on the number of compromised layers}

\begin{figure}[h]
    \centering
    \includegraphics[width=0.8\linewidth]{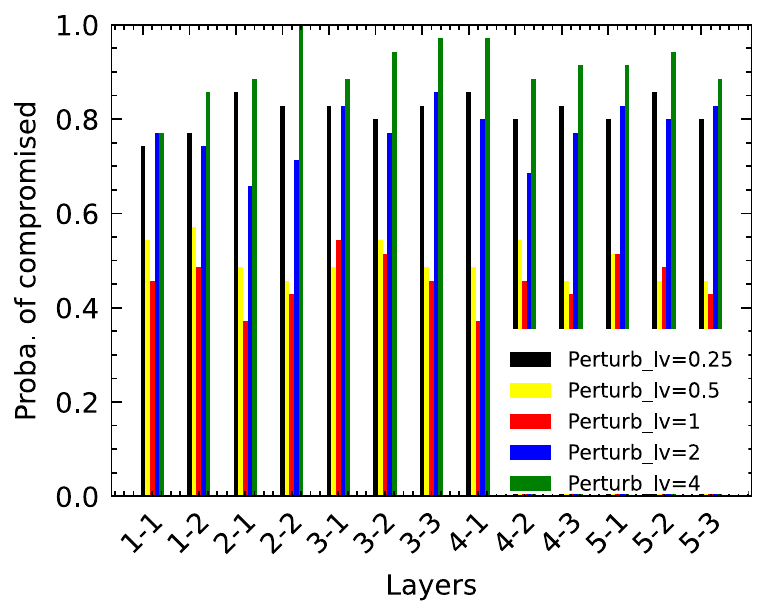}
    \caption{Compromise probability of convolutional layers in the VGG-16 network.}
    \label{figVulnerBarplot}
\end{figure}

\begin{figure}
    \centering
    \includegraphics[width=0.8\linewidth]{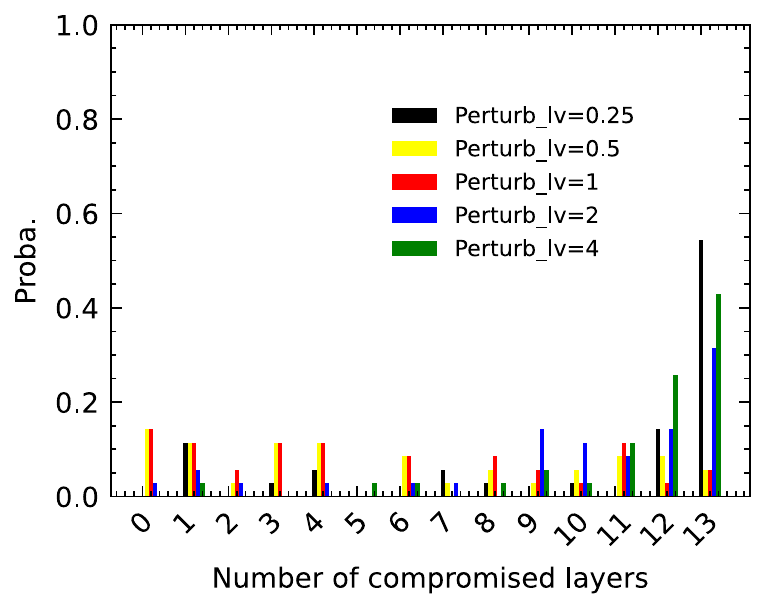}
    \caption{Distribution of the number of compromised layers by single adversarial example.}
    \label{figCompBarplot}
\end{figure}

We also discover that the VGG-16 network can be misled by compromising only a few intermediate layers. We analyze the probability of compromise for each network layer and derive Figure~\ref{figVulnerBarplot}. The statistical distribution of the number of compromised layers with all adversarial examples is shown in Figure~\ref{figCompBarplot}. 

From the previous discussions, perturbations driven by Gaussian noise can cause behavior drift but not necessarily mislead the classifier. As mentioned in Section \ref{sectBDD}, we define a threshold for each convolutional layer in different perturbation levels, and this threshold equals the median value of the behavior deviations when this layer processes the input image with pure Gaussian noise. Intuitively, a layer is called compromised when its behavior under adversarial examples deviates further than it should be under the same input mixed with Gaussian noise.

As shown in Figure~\ref{figVulnerBarplot}, the adversarial examples with the lowest attack strength have approximately 40\% of the chance to compromise any layer. This probability increases significantly and reaches 80\% when we increase the attack strength to be greater than 2 or less than 0.25.

In Figure~\ref{figCompBarplot}, we count the number of convolutional layers that have been compromised for all adversarial examples. We discover that there are some ultimate adversarial examples that can compromise all convolutional layers in the VGG-16 network. This chance of getting the ultimate adversarial examples increases when the attack strength is greater than 2 or less than 0.25. In addition, adversarial samples exist that can cause misclassification by compromising zero or only a small number of layers.

\section{Conclusion}
In this work, we generate adversarial examples using DLFuzz. Through Grad-CAM, we were able to analyze the decision-making procedure of the VGG-16 network layer by layer. In doing so, we were able to show that compared to Gaussian random noise, intentionally
generated adversarial perturbations cause more severe behavioral deviations. Furthermore, we were able to show that in many cases, only a few intermediate blocks of a DNN need to be compromised
in order to manipulate the final decision. Finally, we demonstrate that, in particular, the layers $Block4\_conv1$ and $Block5\_cov1$ of the VGG-16 model are more susceptible to adversarial attacks.

\section{Future Work}
Utilizing our proposed approach, it is possible to find vulnerable layers in a DNN model. In future work, we want to explore the use of a zero-bias layer \cite{9679426} to create more robust DNNs. In addition, it would be interesting to investigate how the focal point changes during the decision-making process of the DNN, and to what extent this correlates with the deliberate perturbations. 
\label{sectCC}

\section*{Acknowledgment}

This research was supported  in  part  by  the  Air  Force  Research  Laboratory Information Directorate,  through  the  Air  Force  Office  of  Scientific  Research  Summer  Faculty Fellowship Program®, Contract Numbers FA8750-15-3-6003, FA9550-15-0001 and FA9550-20-F-0005. This research was also partially supported by the National Science Foundation under Grant No. 2150213.

\bibliographystyle{IEEEtran}
\bibliography{IPCCC2022.bib}

\begin{thebibliography}{10}
\providecommand{\url}[1]{#1}
\csname url@samestyle\endcsname
\providecommand{\newblock}{\relax}
\providecommand{\bibinfo}[2]{#2}
\providecommand{\BIBentrySTDinterwordspacing}{\spaceskip=0pt\relax}
\providecommand{\BIBentryALTinterwordstretchfactor}{4}
\providecommand{\BIBentryALTinterwordspacing}{\spaceskip=\fontdimen2\font plus
\BIBentryALTinterwordstretchfactor\fontdimen3\font minus
  \fontdimen4\font\relax}
\providecommand{\BIBforeignlanguage}[2]{{%
\expandafter\ifx\csname l@#1\endcsname\relax
\typeout{** WARNING: IEEEtran.bst: No hyphenation pattern has been}%
\typeout{** loaded for the language `#1'. Using the pattern for}%
\typeout{** the default language instead.}%
\else
\language=\csname l@#1\endcsname
\fi
#2}}
\providecommand{\BIBdecl}{\relax}
\BIBdecl

\bibitem{simonyan2014very}
K.~Simonyan and A.~Zisserman, ``Very deep convolutional networks for
  large-scale image recognition,'' \emph{arXiv preprint arXiv:1409.1556}, 2014.

\bibitem{10.1145/3132747.3132785}
\BIBentryALTinterwordspacing
K.~Pei, Y.~Cao, J.~Yang, and S.~Jana, ``Deepxplore: Automated whitebox testing
  of deep learning systems,'' in \emph{Proceedings of the 26th Symposium on
  Operating Systems Principles}, ser. SOSP '17.\hskip 1em plus 0.5em minus
  0.4em\relax New York, NY, USA: Association for Computing Machinery, 2017, p.
  1–18. [Online]. Available: \url{https://doi.org/10.1145/3132747.3132785}
\BIBentrySTDinterwordspacing

\bibitem{miller2020adversarial}
D.~J. Miller, Z.~Xiang, and G.~Kesidis, ``Adversarial learning targeting deep
  neural network classification: A comprehensive review of defenses against
  attacks,'' \emph{Proceedings of the IEEE}, vol. 108, no.~3, pp. 402--433,
  2020.

\bibitem{luo2018towards}
B.~Luo, Y.~Liu, L.~Wei, and Q.~Xu, ``Towards imperceptible and robust
  adversarial example attacks against neural networks,'' in \emph{Proceedings
  of the AAAI Conference on Artificial Intelligence}, vol.~32, no.~1, 2018.

\bibitem{englert2019machine}
C.~Englert, P.~Galler, P.~Harris, and M.~Spannowsky, ``Machine learning
  uncertainties with adversarial neural networks,'' \emph{The European Physical
  Journal C}, vol.~79, no.~1, pp. 1--10, 2019.

\bibitem{ignatiev2020towards}
A.~Ignatiev, ``Towards trustable explainable ai.'' in \emph{IJCAI}, 2020, pp.
  5154--5158.

\bibitem{wu2021wider}
B.~Wu, J.~Chen, D.~Cai, X.~He, and Q.~Gu, ``Do wider neural networks really
  help adversarial robustness?'' \emph{Advances in Neural Information
  Processing Systems}, vol.~34, pp. 7054--7067, 2021.

\bibitem{gong2021maxup}
C.~Gong, T.~Ren, M.~Ye, and Q.~Liu, ``Maxup: Lightweight adversarial training
  with data augmentation improves neural network training,'' in
  \emph{Proceedings of the IEEE/CVF Conference on Computer Vision and Pattern
  Recognition}, 2021, pp. 2474--2483.

\bibitem{gao2019convergence}
R.~Gao, T.~Cai, H.~Li, C.-J. Hsieh, L.~Wang, and J.~D. Lee, ``Convergence of
  adversarial training in overparametrized neural networks,'' \emph{Advances in
  Neural Information Processing Systems}, vol.~32, 2019.

\bibitem{9099600}
J.~Guo, Y.~Zhao, H.~Song, and Y.~Jiang, ``Coverage guided differential
  adversarial testing of deep learning systems,'' \emph{IEEE Transactions on
  Network Science and Engineering}, vol.~8, no.~2, pp. 933--942, 2021.

\bibitem{xie2019deephunter}
X.~Xie, L.~Ma, F.~Juefei-Xu, M.~Xue, H.~Chen, Y.~Liu, J.~Zhao, B.~Li, J.~Yin,
  and S.~See, ``Deephunter: a coverage-guided fuzz testing framework for deep
  neural networks,'' in \emph{Proceedings of the 28th ACM SIGSOFT International
  Symposium on Software Testing and Analysis}, 2019, pp. 146--157.

\bibitem{odena2019tensorfuzz}
A.~Odena, C.~Olsson, D.~Andersen, and I.~Goodfellow, ``Tensorfuzz: Debugging
  neural networks with coverage-guided fuzzing,'' in \emph{International
  Conference on Machine Learning}.\hskip 1em plus 0.5em minus 0.4em\relax PMLR,
  2019, pp. 4901--4911.

\bibitem{yang2022revisiting}
Z.~Yang, J.~Shi, M.~H. Asyrofi, and D.~Lo, ``Revisiting neuron coverage metrics
  and quality of deep neural networks,'' \emph{arXiv preprint
  arXiv:2201.00191}, 2022.

\bibitem{harel2020neuron}
F.~Harel-Canada, L.~Wang, M.~A. Gulzar, Q.~Gu, and M.~Kim, ``Is neuron coverage
  a meaningful measure for testing deep neural networks?'' in \emph{Proceedings
  of the 28th ACM Joint Meeting on European Software Engineering Conference and
  Symposium on the Foundations of Software Engineering}, 2020, pp. 851--862.

\bibitem{lamb2019interpolated}
A.~Lamb, V.~Verma, J.~Kannala, and Y.~Bengio, ``Interpolated adversarial
  training: Achieving robust neural networks without sacrificing too much
  accuracy,'' in \emph{Proceedings of the 12th ACM Workshop on Artificial
  Intelligence and Security}, 2019, pp. 95--103.

\bibitem{samangouei2018defense}
P.~Samangouei, M.~Kabkab, and R.~Chellappa, ``Defense-gan: Protecting
  classifiers against adversarial attacks using generative models,''
  \emph{arXiv preprint arXiv:1805.06605}, 2018.

\bibitem{zheng2018robust}
Z.~Zheng and P.~Hong, ``Robust detection of adversarial attacks by modeling the
  intrinsic properties of deep neural networks,'' \emph{Advances in Neural
  Information Processing Systems}, vol.~31, 2018.

\bibitem{gondara2016medical}
L.~Gondara, ``Medical image denoising using convolutional denoising
  autoencoders,'' in \emph{2016 IEEE 16th international conference on data
  mining workshops (ICDMW)}.\hskip 1em plus 0.5em minus 0.4em\relax IEEE, 2016,
  pp. 241--246.

\bibitem{gu2014towards}
S.~Gu and L.~Rigazio, ``Towards deep neural network architectures robust to
  adversarial examples,'' \emph{arXiv preprint arXiv:1412.5068}, 2014.

\bibitem{yadav2022integrated}
A.~Yadav, A.~Upadhyay, and S.~Sharanya, ``An integrated auto encoder-block
  switching defense approach to prevent adversarial attacks,'' \emph{arXiv
  preprint arXiv:2203.10930}, 2022.

\bibitem{hwang2019puvae}
U.~Hwang, J.~Park, H.~Jang, S.~Yoon, and N.~I. Cho, ``Puvae: A variational
  autoencoder to purify adversarial examples,'' \emph{IEEE Access}, vol.~7, pp.
  126\,582--126\,593, 2019.

\bibitem{5206848}
J.~Deng, W.~Dong, R.~Socher, L.-J. Li, K.~Li, and L.~Fei-Fei, ``Imagenet: A
  large-scale hierarchical image database,'' in \emph{2009 IEEE Conference on
  Computer Vision and Pattern Recognition}, 2009, pp. 248--255.

\bibitem{Selvaraju_2017_ICCV}
R.~R. Selvaraju, M.~Cogswell, A.~Das, R.~Vedantam, D.~Parikh, and D.~Batra,
  ``Grad-cam: Visual explanations from deep networks via gradient-based
  localization,'' in \emph{Proceedings of the IEEE International Conference on
  Computer Vision (ICCV)}, Oct 2017.

\bibitem{8591457}
D.~L. Marino, C.~S. Wickramasinghe, and M.~Manic, ``An adversarial approach for
  explainable ai in intrusion detection systems,'' in \emph{IECON 2018 - 44th
  Annual Conference of the IEEE Industrial Electronics Society}, 2018, pp.
  3237--3243.

\bibitem{selvaraju2017grad}
R.~R. Selvaraju, M.~Cogswell, A.~Das, R.~Vedantam, D.~Parikh, and D.~Batra,
  ``Grad-cam: Visual explanations from deep networks via gradient-based
  localization,'' in \emph{Proceedings of the IEEE international conference on
  computer vision}, 2017, pp. 618--626.

\bibitem{ribeiro2016should}
M.~T. Ribeiro, S.~Singh, and C.~Guestrin, ``" why should i trust you?"
  explaining the predictions of any classifier,'' in \emph{Proceedings of the
  22nd ACM SIGKDD international conference on knowledge discovery and data
  mining}, 2016, pp. 1135--1144.

\bibitem{lsa}
\BIBentryALTinterwordspacing
M.~Khalooei, M.~M. Homayounpour, and M.~Amirmazlaghani, ``Layer-wise
  regularized adversarial training using layers sustainability analysis (lsa)
  framework,'' 2022. [Online]. Available:
  \url{https://arxiv.org/abs/2202.02626}
\BIBentrySTDinterwordspacing

\bibitem{Haizhong}
\BIBentryALTinterwordspacing
H.~Zheng, Z.~Zhang, H.~Lee, and A.~Prakash, ``Understanding and diagnosing
  vulnerability under adversarial attacks,'' 2020. [Online]. Available:
  \url{https://arxiv.org/abs/2007.08716}
\BIBentrySTDinterwordspacing

\bibitem{goodfellow2014explaining}
I.~J. Goodfellow, J.~Shlens, and C.~Szegedy, ``Explaining and harnessing
  adversarial examples,'' \emph{arXiv preprint arXiv:1412.6572}, 2014.

\bibitem{cian2020evaluating}
D.~Cian, J.~van Gemert, and A.~Lengyel, ``Evaluating the performance of the
  lime and grad-cam explanation methods on a lego multi-label image
  classification task,'' \emph{arXiv preprint arXiv:2008.01584}, 2020.

\bibitem{2008.02312}
\BIBentryALTinterwordspacing
R.~Fu, Q.~Hu, X.~Dong, Y.~Guo, Y.~Gao, and B.~Li, ``Axiom-based grad-cam:
  Towards accurate visualization and explanation of cnns,'' 2020. [Online].
  Available: \url{https://arxiv.org/abs/2008.02312}
\BIBentrySTDinterwordspacing

\bibitem{poerner-etal-2018-evaluating}
\BIBentryALTinterwordspacing
N.~Poerner, H.~Sch{\"u}tze, and B.~Roth, ``Evaluating neural network
  explanation methods using hybrid documents and morphosyntactic agreement,''
  in \emph{Proceedings of the 56th Annual Meeting of the Association for
  Computational Linguistics (Volume 1: Long Papers)}.\hskip 1em plus 0.5em
  minus 0.4em\relax Melbourne, Australia: Association for Computational
  Linguistics, Jul. 2018, pp. 340--350. [Online]. Available:
  \url{https://aclanthology.org/P18-1032}
\BIBentrySTDinterwordspacing

\bibitem{app112110417}
\BIBentryALTinterwordspacing
F.~Gabbay, S.~Bar-Lev, O.~Montano, and N.~Hadad, ``A lime-based explainable
  machine learning model for predicting the severity level of covid-19
  diagnosed patients,'' \emph{Applied Sciences}, vol.~11, no.~21, 2021.
  [Online]. Available: \url{https://www.mdpi.com/2076-3417/11/21/10417}
\BIBentrySTDinterwordspacing

\bibitem{ILSVRC15}
O.~Russakovsky, J.~Deng, H.~Su, J.~Krause, S.~Satheesh, S.~Ma, Z.~Huang,
  A.~Karpathy, A.~Khosla, M.~Bernstein, A.~C. Berg, and L.~Fei-Fei, ``{ImageNet
  Large Scale Visual Recognition Challenge},'' \emph{International Journal of
  Computer Vision (IJCV)}, vol. 115, no.~3, pp. 211--252, 2015.

\bibitem{9679426}
Y.~Liu, Y.~Chen, J.~Wang, S.~Niu, D.~Liu, and H.~Song, ``Zero-bias deep neural
  network for quickest rf signal surveillance,'' in \emph{2021 IEEE
  International Performance, Computing, and Communications Conference (IPCCC)},
  2021, pp. 1--8.

\end{thebibliography}

\end{document}